\documentclass[a4wide,twocolumn,9pt]{extarticle}
\usepackage[a4paper,top=0.85in,left=0.75in,bottom=1in,right=0.52in]{geometry}
\usepackage{amssymb}
\usepackage{amsmath}
\usepackage{amsfonts}
\usepackage{graphicx}
\usepackage{multirow}
\usepackage{color}
\usepackage{algorithm}
\usepackage{algorithmic}
\usepackage{mathrsfs}
\usepackage[svgnames]{xcolor}
\usepackage{url}
\usepackage{hyperref}
\usepackage{soul} 
\usepackage{caption}
\DeclareMathOperator*{\argmax}{argmax}

\newcommand{\surl}[1]
{
	\urlstyle{same}\url{#1}
}

\newcommand{\captionmoveup}{\vspace{-0.0in}}
\begin{document}

\clubpenalty=10000
\widowpenalty = 10000

\title{\textsf{\textbf{Learning Points and Routes to Recommend Trajectories}}}

\pdfinfo{
/Title (Learning Points and Routes to Recommend Trajectories)
/Author (Dawei Chen, Cheng Soon Ong, Lexing Xie)
} 

\author{
    \textsf{Dawei Chen$^{*\dagger}$,~Cheng Soon Ong$^{\dagger *}$,~Lexing Xie$^{*\dagger}$}\\
    \textsf{$^*$The Australian National University,~$^\dagger$Data 61, CSIRO, Australia}\\
    \textsf{\{u5708856, chengsoon.ong, lexing.xie\}@anu.edu.au}
}

\date{}

\maketitle

\begin{abstract}
The problem of recommending tours to travellers is an important and broadly studied area.
Suggested solutions include various approaches of points-of-interest (POI)
recommendation and route planning.
We consider the task of recommending a sequence of POIs,
that simultaneously uses information about POIs and routes.
Our approach unifies the treatment of various sources of information
by representing them as features in machine learning algorithms, enabling us to learn from past behaviour. 
Information about POIs are used to learn a POI ranking model that accounts for the start and end points of tours.
Data about previous trajectories are used for learning transition patterns between POIs that enable us to recommend probable routes.
In addition, a probabilistic model is proposed to combine the results of POI ranking and the POI to POI transitions.
We propose a new F$_1$ score on pairs of POIs that capture the order of visits.
Empirical results show that our approach improves on recent methods, 
and demonstrate that combining points and routes enables better trajectory recommendations.
\end{abstract}

\section*{Keywords}
Trajectory recommendation; learning to rank; planning

\section{Introduction}
\label{sec:intro}

This paper proposes a novel solution to recommend travel routes in cities.
A large amount of location traces are becoming available from ubiquitous location tracking devices.
For example, FourSquare has 50 million monthly users who have made 8 billion check-ins~\cite{4sq},
and Flickr hosts over 2 billion geo-tagged public photos~\cite{flickr}.
This growing trend in rich geolocation data
provide new opportunities for better
travel planning traditionally done with written travel guides.
Good solutions to these problems will in turn lead to better urban experiences for residents and visitors alike, and foster sharing of even more location-based behavioural data.

\begin{figure}[t]
	\centering
	\includegraphics[width=\columnwidth]{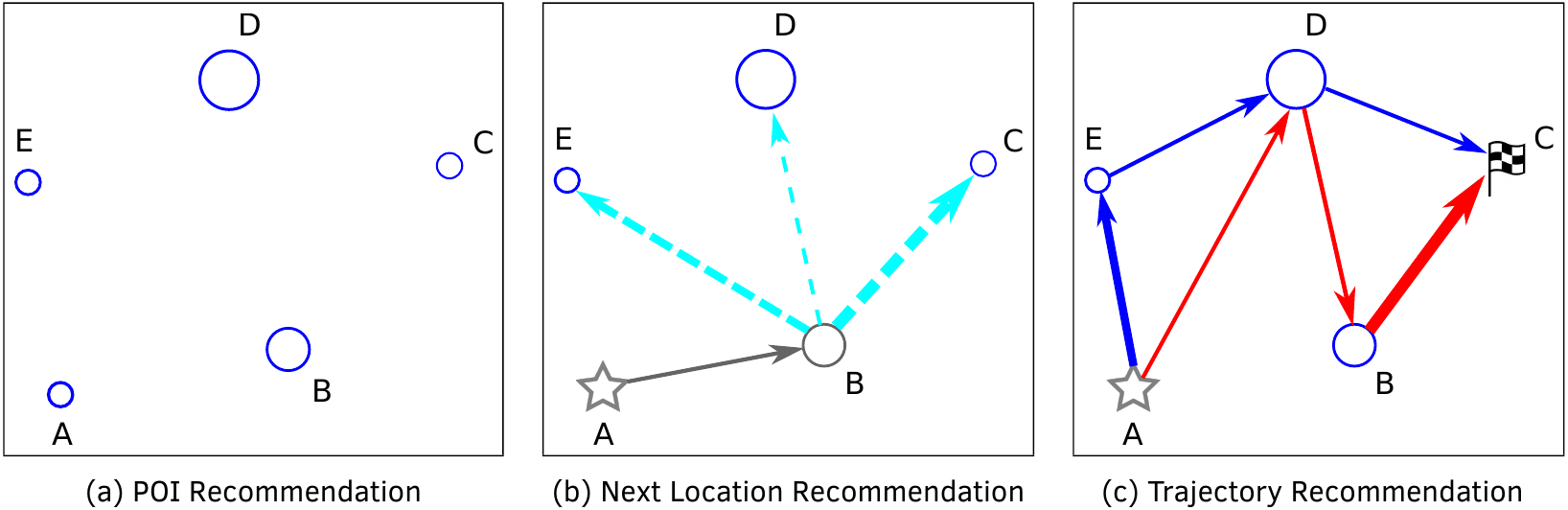}
	\caption{Three settings of trajectory recommendation problems.
Node size: POI score; edge width: transition score between pairs of POIs;
grey: observed;
star: starting location; flag: ending location. See Section~\ref{sec:intro} for details.
}
	\label{fig:threesettings}\captionmoveup
\end{figure}

There are several settings of recommendation problems for locations and routes, as illustrated in Figure~\ref{fig:threesettings}.
We summarise recent work most related to formulating and solving learning problems on assembling routes from POIs,
and refer the reader to a number of recent surveys~\cite{bao2015recommendations,zheng2015trajectory,zheng2014urban} for general overviews of the area.
The first setting can be called POI recommendation (Figure~\ref{fig:threesettings}(a)). Each location (A to E) is scored with geographic and behavioural information such as category, reviews, popularity, spatial information such as distance, and temporal information such as travel time uncertainty, time of the day or day of the week.
A popular approach is to recommend POIs with a collaborative filtering model
on user-location affinity~\cite{shi2011personalized}, with additional ways to incorporate spatial~\cite{lian2014geomf,liu2014exploiting}, temporal~\cite{yuan2013timeaware,hsieh2014mining,gao2013temporal}, or spatial-temporal~\cite{yuan2014graph} information.

Figure~\ref{fig:threesettings}(b) illustrates the second setting: next location recommendation.
Here the input is a partial trajectory (e.g. started at point A and currently at point B), the task of the algorithm is to score the next candidate location (e.g, C, D and E) based on the perceived POI score and transition compatibility with input $A\rightarrow B$.
It is a variant of POI recommendation except both the user and locations travelled to date are given. The solutions to this problem include incorporating Markov chains into collaborative filtering~\cite{fpmc10,ijcai13,zhang2015location},
quantifying tourist traffic flow between points-of-interest~\cite{zheng2012patterns},
formulating a binary decision or ranking problem~\cite{baraglia2013learnext}, and predict the next location with sequence models such as recurrent neural networks~\cite{aaai16}.

This paper considers the final setting: trajectory recommendation (Figure~\ref{fig:threesettings}(c)). Here the input are some factors about the desired route, e.g. starting point A and end point C, along with auxiliary information such as the desired length of trip. The algorithm needs to take into account location desirability (as indicated by node size) and transition compatibility (as indicated by edge width), and compare route hypotheses such as A-D-B-C and A-E-D-C. Existing work in this area either uses heuristic combination of locations and routes~\cite{lu2010photo2trip,ijcai15,lu2012personalized}, or formulates an optimisation problem that is not informed or evaluated by behaviour history~\cite{gioniswsdm14,chen2015tripplanner}.
We note, however, that two desired qualities are still
missing from the current solutions to trajectory recommendation.
The first is a principled method to jointly learn POI ranking (a prediction problem)
and optimise for route creation (a planning problem).
The second is a unified way to incorporate various features
such as location, time, distance, user profile and social interactions,
as they tend to get specialised and separate treatments.
This work aims to address both challenges. 
We propose a novel way to learn point preferences and routes jointly.
In Section~\ref{sec:feature}, we describe the features that are used to ranking points,
and POI to POI transitions that are factorised along 
different types of location properties.
Section~\ref{sec:recommendation} details a number of our proposed approaches to recommend trajectories.
We evaluate the proposed algorithms on trajectories from five different cities in Section~\ref{sec:experiment}.
The main contributions of this work are:
\begin{itemize}
\setlength{\itemsep}{-2pt}
\item We propose a novel algorithm to jointly optimise point preferences and routes. We find that learning-based approaches generally outperform heuristic route recommendation~\cite{ijcai15}.
Incorporating transitions to POI ranking results in a better sequence of POIs, and avoiding sub-tours further improves performance of classical Markov chain methods.
\item Our approach is feature-driven and learns from past behaviour without having to design specialised treatment for spatial, temporal or social information. It incorporates information about location, POI categories and behaviour history, and can use additional time, user, or social information if available.
\item We show good performance compared to recent results~\cite{ijcai15}, and also quantify the contributions from different components, such as ranking points, scoring transitions, and routing.
\item We propose a new metric to evaluate trajectories, pairs-F$_1$, to capture the order in which POIs are visited. Pairs-F$_1$ lies between 0 and 1, and achieves 1 if and only if the recommended trajectory is exactly the same as the ground truth.
\end{itemize}
Supplemental material, benchmark data and results are available online at \surl{https://bitbucket.org/d-chen/tour-cikm16}.

\section{POI, Query and Transition}
\label{sec:feature}

The goal of tour recommendation is to suggest a sequence of POIs, $(p_1, \ldots, p_L)$, of length $L$ such that the user's utility is maximised. The user provides the desired start ($p_1=p_s$) and end point ($p_L=p_e$), as well as the number $L$ of POIs desired, from which we propose a trajectory through the city.
The training data consists of a set of tours of varying length in a particular city.
We consider only POIs that have been visited by at least one user in the past, and
construct a graph with POIs as nodes and directed edges representing the observed transitions between pairs of POIs in tours.

We extract the category, popularity (number of distinct visitors)~\cite{ht10}, total number of visits and average visit duration for each POI.
POIs are grouped into $5$ clusters using K-means according to their geographical locations to reflect their neighbourhood.
Furthermore, since we are constrained by the fact that trajectories have to be of length $L$ and start and end at certain points, we hope to improve the recommendation by using this information.
In other words, we use the \textit{query} $q = (p_s, p_e, L)$ to construct new features by contrasting candidate POIs with $p_s$ and $p_e$.
For each of the POI features (category, neighbourhood, popularity, total visits and average duration),
we construct two new features by taking the difference of the feature in POI $p$ with $p_s$ and $p_e$ respectively.
For the category (and neighbourhood), we set the feature to $1$ when their categories (and cluster identities) are the same and $-1$ otherwise.
For popularity, total visits and average duration, we take the real valued difference.
Lastly, we compute the distance from POI $p$ to $p_s$ (and $p_e$) using the Haversine formula~\cite{haversine},
and also include the required length $L$.

\begin{figure}[t]
\includegraphics[width=\columnwidth]{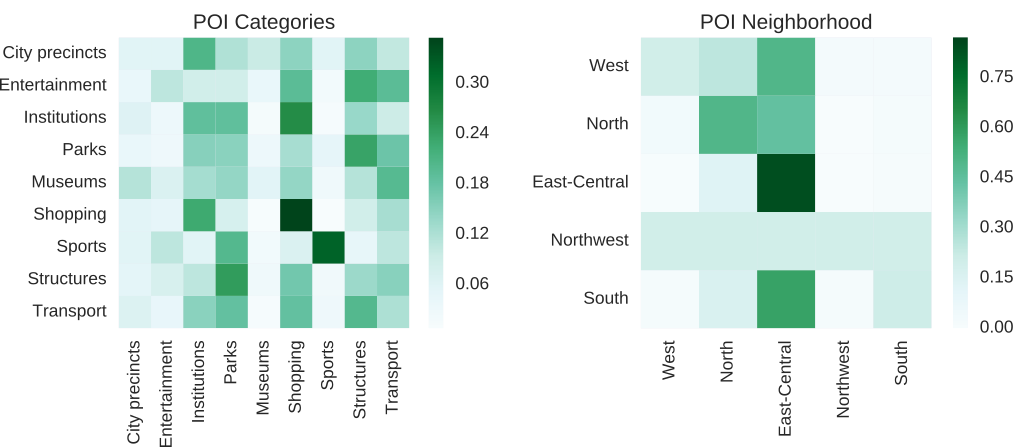}
\caption{Transition matrices for two POI features from Melbourne: POI category and neighbourhood.
}
\label{fig:transmat}\captionmoveup
\end{figure}

In addition to information about each individual POI, a tour recommendation system would benefit
from capturing the likelihood of going from one POI to another different POI. One option would be to
directly model the probability of going from any POI to any other POI, but this has several weaknesses:
Such a model would be unable to handle a new POI (one that has not yet been visited),
or pairs of existing POIs that do not have an observed transition.
Furthermore, even if we restrict ourselves to known POIs and transitions,
there may be locations which are rarely visited,
leading to significant challenges in estimating the probabilities from empirical data.

We model POI transitions using a Markov chain with discrete 
states by factorising the transition probability ($p_i$ to $p_j$) 
as a product of transition probabilities between pairs of individual POI features, 
assuming independence between these feature-wise transitions.
The popularity, total visits and average duration are discretised by binning
them uniformly into $5$ intervals on the log scale.
These feature-to-feature transitions are estimated from data using maximum likelihood principle.
The POI-POI transition probabilities can be efficiently computed by taking the Kronecker product of 
transition matrices for the individual features,
and then updating it based on three additional constraints as well as appropriate normalisation.
First we disallow self-loops by setting the probability of ($p_i$ to $p_i$) to zero.
Secondly, when multiple POIs have identical (discretised) features, we distribute the probability uniformly among POIs in the group.
Third, we remove feature combinations that has no POI in dataset. 
Figure~\ref{fig:transmat} visualises the transition matrices for two POI features, category and neighbourhood, in Melbourne.

\section{Tour Recommendation}
\label{sec:recommendation}

In this section, we first describe the recommendation of points and routes,
then we discuss how to combine them, and finally we propose a method to avoid sub-tours.

\subsection{POI Ranking and Route Planning}
\label{sec:rankplan}

A naive approach would be to recommend trajectories based on the popularity of POIs only,
that is we always suggest the top-$k$ most popular POIs for all visitors given the start and end location.
We call this baseline approach \textsc{PoiPopularity},
and its only adaptation to a particular query is to adjust $k$ to match the desired length.

On the other hand, we can leverage the whole set of POI features described in Section~\ref{sec:feature}
to learn a ranking of POIs using rankSVM, with linear kernel and L$2$ loss~\cite{lranksvm},
\begin{equation*}
\min_{\mathbf{w}} \frac{1}{2}
                  \mathbf{w}^T \mathbf{w} +
                  \underset{p_i, p_j \in \mathcal{P},~ q \in \mathcal{Q}}{C ~\sum}
                  \max \left( 0,~ 1 - \mathbf{w}^T (\phi_{i,q} - \phi_{j,q}) \right)^2,
\end{equation*}
where $\mathbf{w}$ is the parameter vector,
$C > 0$ is a regularisation constant,
$\mathcal{P}$ is the set of POIs to rank,
$\mathcal{Q}$ denotes the queries corresponding to trajectories in training set,
and $\phi_{i,q}$ is the feature vector for POI $p_i$ with respect to query $q$. 
The ranking score of $p_i$ given query $q$ is computed as $R_{i,q} =\mathbf{w}^T \phi_{i,q}$. 

For training the rankSVM, the labels are generated using the number of occurrences of
POI $p$ in trajectories grouped by query $(p_s, p_e, L)$,
without counting the occurrence of $p$ when it is the origin or destination of a trajectory.
Our algorithm, \textsc{PoiRank}, recommends a trajectory for a particular query
by first ranking POIs then takes the top ranked $L-2$ POIs and connects them according to the ranks.

In addition to recommend trajectory by ranking POIs, we can leverage the POI-POI transition probabilities and
recommend a trajectory (with respect to a query) by maximising the transition likelihood.
The maximum likelihood of the Markov chain of transitions is found using a variant of the Viterbi algorithm (with uniform emission probabilities).
We call this approach that only uses the transition probabilities between POIs as \textsc{Markov}.

\subsection{Combine Ranking and Transition}
\label{sec:rank+markov}

We would like to leverage both point ranking and transitions,
i.e., recommending a trajectory that maximises the points ranking of its POIs as well as its transition likelihood at the same time.
To begin with, we transform the ranking scores $R_{j,q}$ of POI $p_j$ with respect to query $q$
to a probability distribution using the softmax function,
\begin{equation}
\label{eq:rankprob}
P_R(p_j | q) = \frac{\exp(R_{j,q})}{\sum_i \exp(R_{i,q})},
\end{equation}
One option to find a trajectory that simultaneously maximises the ranking probabilities of its POIs and its transition likelihood is to optimise the following objective:
\vspace{-0.3em}
\begin{equation*}
    \argmax_{\mathcal{T} \in \mathcal{P}^L} ~\alpha \sum_{k=2}^{L} \log P_R(p_{k} | q) +
                                     (1-\alpha) \sum_{k=1}^{L-1} \log P(p_{k+1} | p_{k}),
\end{equation*}
such that
$p_{1} = p_s, ~ p_{L} = p_e$ and
$p_{k} \in \mathcal{P}, ~1 \le k \le L$.
The first term captures the POI ranking, and the second one incorporates the transition probabilities.
$\mathcal{T} = (p_{1}, \dots, p_{L})$ is any possible trajectory,
$\alpha \in [0, 1]$ is a parameter to trade-off the importance between point ranking and transition,
and can be tuned using cross validation in practice.
Let $S(p; p', q)$ be a convex combination of point ranking and transition,
\vspace{-0.3em}
\begin{equation}\label{eq:combined-score}
    S(p; p', q)  = \alpha \log P_R(p|q) + (1-\alpha) \log P(p|p'),
\end{equation}
then the best path (or walk) can be found using the Viterbi algorithm.
We call this approach that uses both the point ranking and transitions \textsc{Rank+Markov},
with pseudo code shown in Algorithm~\ref{alg:rank+markov},
where $A$ is the score matrix, and entry $A[l, p]$ stores the maximum value associated with the (partial) trajectory
that starts at $p_s$ and ends at $p$ with $l$ POI visits;
$B$ is the backtracking-point matrix, and entry $B[l, p]$ stores the predecessor of $p$ in that (partial) trajectory.
The maximum objective value is $A[L, p_e]$,
and the corresponding trajectory can be found by tracing back from $B[L, p_e]$.

\setlength{\textfloatsep}{0.5em} 

\begin{algorithm}[t]
\caption{\textsc{Rank+Markov}: recommend trajectory with POI ranking and transition}
\label{alg:rank+markov}
\begin{algorithmic}[1]
\STATE \textbf{Input}: $\mathcal{P}, p_s, p_e, L$
\STATE \textbf{Output}: Trajectory $\mathcal{T} = (p_s, \cdots, p_e)$ with $L$ POIs
\STATE Initialise score matrix $A$ and backtracking pointers $B$
\FOR{$p \in \mathcal{P}$}
    \STATE $A[2, p] = S(p; p_s, q)$
    \STATE $B[2, p] = p_s$
\ENDFOR
\FOR{$l=2$ to $L-1$}
    \FOR{$p \in \mathcal{P}$}
        \STATE $A[l+1, p]   = \max_{p' \in \mathcal{P}} \{ A[l, p'] + S(p; p', q) \}$ \label{eq:max}
        \STATE $B[l+1, p]   = \argmax_{p' \in \mathcal{P}} \{ A[l, p'] + S(p; p', q) \}$ \label{eq:argmax}
    \ENDFOR
\ENDFOR
\STATE $\mathcal{T}= \{p_e\}$, $l = L$, $p = p_e$
\REPEAT
    \STATE Prepend $B[l, p]$ to $\mathcal{T}$
    \STATE $l = l - 1$, $p = B[l, p]$
\UNTIL{$l < 2$}
\RETURN $\mathcal{T}$
\end{algorithmic}
\end{algorithm}

\subsection{Avoiding sub-tours} 
\label{sec:nosubtour}

Trajectories recommended by \textsc{Markov} (Section~\ref{sec:rankplan}) and \textsc{Rank+Markov} (Section~\ref{sec:rank+markov})
are found using the maximum likelihood approach, and may contain multiple visits to the same POI.
This is because the best solution from Viterbi decoding may have
circular sub-tours (where a POI already visited earlier in the tour is visited again).
We propose a method for eliminating sub-tours by finding the best path using an integer linear program (ILP),
with sub-tour elimination constraints adapted from the Travelling Salesman Problem~\cite{opt98}.
In particular, given a set of POIs $\mathcal{P}$, the POI-POI transition matrix and a query $q = (p_s, p_e, L)$,
we recommend a trajectory by solving the following ILP:
\vspace{-0.3em}
\begin{alignat}{5}
& \max_{x,u}  ~&& \sum_{i=1}^{N-1} \sum_{j=2}^N ~x_{ij} ~\log P(p_j | p_i)                                                \nonumber \\
& ~s.t. ~&& x_{ij} \in \{0, 1\}, ~x_{ii} = 0, ~u_i \in \mathbf{Z}, ~\forall i, j = 1, \cdots, N                    \label{eq:cons1} \\
&        && \sum_{j=2}^N x_{1j} = \sum_{i=1}^{N-1} x_{iN} = 1, ~\sum_{i=2}^N x_{i1} = \sum_{j=1}^{N-1} x_{Nj} = 0  \label{eq:cons2} \\
&        && \sum_{i=1}^{N-1} x_{ik} = \sum_{j=2}^N x_{kj} \le 1,   ~\forall k=2, \cdots, N-1                       \label{eq:cons3} \\
&        && \sum_{i=1}^{N-1} \sum_{j=2}^N x_{ij} = L-1,                                                            \label{eq:cons4} \\
&        && u_i - u_j + 1 \le (N-1) (1-x_{ij}),                     \forall i, j = 2, \cdots, N                    \label{eq:cons5}
\end{alignat}
where $N=|\mathcal{P}|$ is the number of available POIs and $x_{ij}$ is a binary decision variable
that determines whether the transition from $p_i$ to $p_j$ is in the resulting trajectory.
For brevity, we arrange the POIs such that $p_1 = p_s$ and $p_N = p_e$.
Firstly, the desired trajectory should start from $p_s$ and end at $p_e$ (Constraint~\ref{eq:cons2}).
In addition, any POI could be visited at most once (Constraint~\ref{eq:cons3}).
Moreover, only $L-1$ transitions between POIs are permitted (Constraint~\ref{eq:cons4}),
i.e., the number of POI visits should be exactly $L$ (including $p_s$ and $p_e$).
The last constraint, where $u_i$ is an auxiliary variable,
enforces that only a single sequence of POIs without sub-tours is permitted in the trajectory.
We solve this ILP using the Gurobi optimisation package~\cite{gurobi}, 
and the resulting trajectory is constructed by tracing the non-zeros in $x$. 
We call our method that uses the POI-POI transition matrix to recommend paths
without circular sub-tours \textsc{MarkovPath}.

Sub-tours in trajectories recommended by \textsc{Rank+Markov} can be eliminated in a similar manner,
we solve an ILP by optimising the following objective 
with the same constraints described above,
\vspace{-1em}
\begin{equation}
\label{eq:obj2}
\max_{x,u} \sum_{i=1}^{N-1} \sum_{j=2}^N ~x_{ij} ~S(p_j; p_i, q),
\end{equation}
where $S(p_j;p_i,q)$ incorporates both point ranking and transition, as defined in Equation~(\ref{eq:combined-score}).
This algorithm is called \textsc{Rank+MarkovPath} in the experiments.

\setlength{\textfloatsep}{2em} 

\section{Experiment on Flickr Photos}
\label{sec:experiment}

\begin{table}[t]
\caption{Statistics of trajectory dataset}
\label{tab:data}
\centering
\begin{tabular}{l*{5}{r}} \hline
\textbf{Dataset} & \textbf{\#Photos} & \textbf{\#Visits} & \textbf{\#Traj.} & \textbf{\#Users} \\ \hline
Edinburgh & 82,060 & 33,944 & 5,028 & 1,454 \\
Glasgow & 29,019 & 11,434 & 2,227 & 601 \\
Melbourne & 94,142 & 23,995 & 5,106 & 1,000 \\
Osaka & 392,420 & 7,747 & 1,115 & 450 \\
Toronto & 157,505 & 39,419 & 6,057 & 1,395 \\
\hline
\end{tabular}\captionmoveup
\end{table}

We evaluate the algorithms above on datasets with trajectories extracted from Flickr photos~\cite{thomee2016yfcc100m} in five cities,
namely, Edinburgh, Glasgow, Melbourne, Osaka and Toronto, with statistics shown in Table~\ref{tab:data}.
The Melbourne dataset is built using approaches proposed in earlier work~\cite{ht10, ijcai15},
and the other four datasets are provided by Lim et al.~\cite{ijcai15}.

We use leave-one-out cross validation to evaluate different trajectory recommendation algorithms,
i.e., when testing on a trajectory, all other trajectories are used for training.
We compare with a number of baseline approaches such as \textsc{Random},
which naively chooses POIs uniformly at random (without replacement) from the set $\mathcal{P} \setminus \{p_s, p_e \}$ to form a trajectory,
and \textsc{PoiPopularity} (Section~\ref{sec:rankplan}), which recommends trajectories based on the popularity of POIs only.
Among the related approaches from recent literature,
\textsc{PersTour}~\cite{ijcai15} explores POI features as well as the sub-tour elimination constraints (Section~\ref{sec:nosubtour}),
with an additional time budget, and its variant \textsc{PersTour-L},
which replaces the time budget with a constraint of trajectory length.
Variants of point-ranking and route-planning approaches including \textsc{PoiRank} and \textsc{Markov} (Section~\ref{sec:rankplan}),
which utilises either POI features or POI-POI transitions,
and \textsc{Rank+Markov} (Section~\ref{sec:rank+markov}) that captures both types of information. 
Variants that employ additional sub-tour elimination constraints
(\textsc{MarkovPath} and \textsc{Rank+MarkovPath}, Section~\ref{sec:nosubtour}) are also included.
A summary of the various trajectory recommendation approaches can be found in Table~\ref{tab:algsummary}.

\begin{table}[t]
\caption{Summary of information captured by different trajectory recommendation algorithms}
\label{tab:algsummary}
\centering
\setlength{\tabcolsep}{3pt} 
\begin{tabular}{l|*{4}{c}} \hline
                                & Query    & POI      & Trans.     & No sub-tours \\ \hline 
\textsc{Random}                 & $\times$ & $\times$ & $\times$   & $\times$     \\ 
\textsc{PersTour}\cite{ijcai15} & $\times$ & $\surd$  & $\times$   & $\surd$      \\ 
\textsc{PersTour-L}             & $\times$ & $\surd$  & $\times$   & $\surd$      \\ 
\textsc{PoiPopularity}          & $\times$ & $\surd$  & $\times$   & $\times$     \\ 
\textsc{PoiRank}                & $\surd$  & $\surd$  & $\times$   & $\times$     \\ 
\textsc{Markov}                 & $\times$ & $\surd$  & $\surd$    & $\times$     \\ 
\textsc{MarkovPath}             & $\times$ & $\surd$  & $\surd$    & $\surd$      \\ 
\textsc{Rank+Markov}            & $\surd$  & $\surd$  & $\surd$    & $\times$     \\ 
\textsc{Rank+MarkovPath}        & $\surd$  & $\surd$  & $\surd$    & $\surd$      \\ 
\hline
\end{tabular}
\end{table}

\begin{table*}[t]
\caption{Performance comparison on five datasets in terms of F$_1$ score. 
        The best method for each dataset (i.e., a column) is shown in bold, the second best is shown in {\em italic}.}
\label{tab:f1}
\centering
\setlength{\tabcolsep}{10pt} 
\begin{tabular}{l|ccccc} \hline
 & Edinburgh & Glasgow & Melbourne & Osaka & Toronto \\ \hline
\textsc{Random} & $0.570\pm0.139$ & $0.632\pm0.123$ & $0.558\pm0.149$ & $0.621\pm0.115$ & $0.621\pm0.129$ \\
\textsc{PersTour}\cite{ijcai15} & $0.656\pm0.223$ & $\mathbf{0.801\pm0.213}$ & $0.483\pm0.208$ & $0.686\pm0.231$ & $0.720\pm0.215$ \\
\textsc{PersTour-L} & $0.651\pm0.143$ & $0.660\pm0.102$ & $0.576\pm0.141$ & $0.686\pm0.137$ & $0.643\pm0.113$ \\
\textsc{PoiPopularity} & $\mathbf{0.701\pm0.160}$ & $0.745\pm0.166$ & $0.620\pm0.136$ & $0.663\pm0.125$ & $0.678\pm0.121$ \\
\textsc{PoiRank} & $\mathit{0.700\pm0.155}$ & $\mathit{0.768\pm0.171}$ & $\mathit{0.637\pm0.142}$ & $\mathbf{0.745\pm0.173}$ & $\mathbf{0.754\pm0.170}$ \\
\textsc{Markov} & $0.645\pm0.169$ & $0.725\pm0.167$ & $0.577\pm0.168$ & $0.697\pm0.150$ & $0.669\pm0.151$ \\
\textsc{MarkovPath} & $0.678\pm0.149$ & $0.732\pm0.168$ & $0.595\pm0.148$ & $0.706\pm0.150$ & $0.688\pm0.138$ \\
\textsc{Rank+Markov} & $0.659\pm0.174$ & $0.754\pm0.173$ & $0.613\pm0.166$ & $0.715\pm0.164$ & $0.723\pm0.185$ \\
\textsc{Rank+MarkovPath} & $0.697\pm0.152$ & $0.762\pm0.167$ & $\mathbf{0.639\pm0.146}$ & $\mathit{0.732\pm0.162}$ & $\mathit{0.751\pm0.170}$ \\
\hline
\end{tabular}
\vspace{-1.2em}
\end{table*}

\begin{table*}[t]
\caption{Performance comparison on five datasets in terms of pairs-F$_1$ score.
        The best method for each dataset (i.e., a column) is shown in bold, the second best is shown in {\em italic}.}
\label{tab:pairf1}
\centering
\setlength{\tabcolsep}{10pt} 
\begin{tabular}{l|ccccc} \hline
 & Edinburgh & Glasgow & Melbourne & Osaka & Toronto \\ \hline
\textsc{Random} & $0.261\pm0.155$ & $0.320\pm0.168$ & $0.248\pm0.147$ & $0.304\pm0.142$ & $0.310\pm0.167$ \\
\textsc{PersTour}\cite{ijcai15} & $0.417\pm0.343$ & $\mathbf{0.643\pm0.366}$ & $0.216\pm0.265$ & $0.468\pm0.376$ & $0.504\pm0.354$ \\
\textsc{PersTour-L} & $0.359\pm0.207$ & $0.352\pm0.162$ & $0.266\pm0.140$ & $0.406\pm0.238$ & $0.333\pm0.163$ \\
\textsc{PoiPopularity} & $\mathit{0.436\pm0.259}$ & $0.507\pm0.298$ & $0.316\pm0.178$ & $0.365\pm0.190$ & $0.384\pm0.201$ \\
\textsc{PoiRank} & $0.432\pm0.251$ & $\mathit{0.548\pm0.311}$ & $0.339\pm0.203$ & $\mathbf{0.511\pm0.309}$ & $\mathbf{0.518\pm0.296}$ \\
\textsc{Markov} & $0.417\pm0.248$ & $0.495\pm0.296$ & $0.288\pm0.195$ & $0.445\pm0.266$ & $0.407\pm0.241$ \\
\textsc{MarkovPath} & $0.400\pm0.235$ & $0.485\pm0.293$ & $0.294\pm0.187$ & $0.442\pm0.260$ & $0.405\pm0.231$ \\
\textsc{Rank+Markov} & $\mathbf{0.444\pm0.263}$ & $0.545\pm0.306$ & $\mathbf{0.351\pm0.220}$ & $0.486\pm0.288$ & $0.512\pm0.303$ \\
\textsc{Rank+MarkovPath} & $0.428\pm0.245$ & $0.533\pm0.303$ & $\mathit{0.344\pm0.206}$ & $\mathit{0.489\pm0.287}$ & $\mathit{0.514\pm0.297}$ \\
\hline
\end{tabular}
\vspace{-1em}
\end{table*}

\subsection{Performance metrics}
\label{sec:metric}

\begin{figure}[t]
	\centering
	\includegraphics[width=\columnwidth]{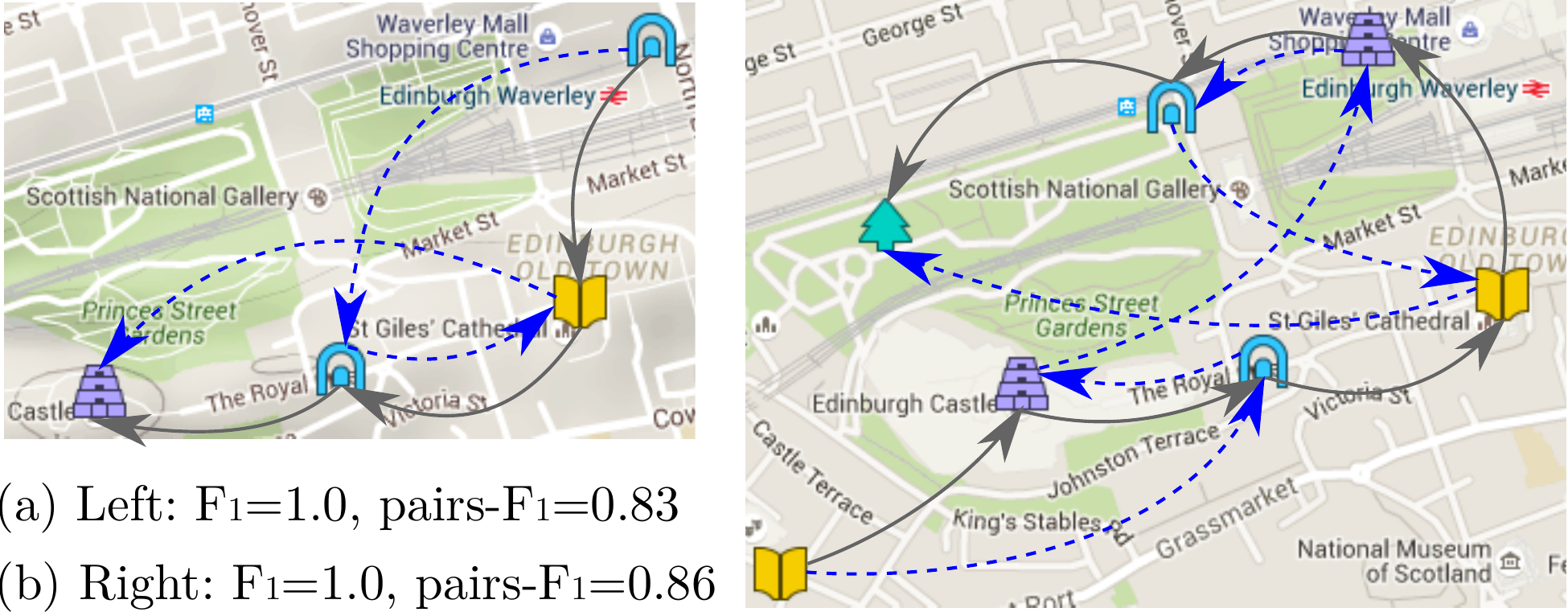}
	\caption{Examples for F$_1$ vs pairs-F$_1$ as evaluation metric.
Solid grey: ground truth; dashed blue: recommended trajectories. See Section~\ref{sec:metric} for details.}
	\label{fig:pairf1}\captionmoveup
\end{figure}

A commonly used metric for evaluating POI and trajectory recommendation is
the F$_1$ score on points, which is the harmonic mean of precision and recall of POIs in trajectory~\cite{ijcai15}.
While being good at measuring whether POIs are correctly recommended,
F$_1$ score on points ignores the visiting order between POIs.
We propose a new metric $\text{pairs-F}_1$ that considers both POI identity and visiting order,
by measuring the F$_1$ score of every pair of POIs,
whether they are adjacent or not in trajectory,
\begin{displaymath}
\text{pairs-F}_1 = \frac{2 P_{\textsc{pair}} R_{\textsc{pair}}}
                        {P_{\textsc{pair}} + R_{\textsc{pair}}},
\end{displaymath}
where $P_{\textsc{pair}}$ and $R_{\textsc{pair}}$ are the precision and recall of ordered POI pairs respectively. 
Pairs-F$_1$ takes values between 0 and 1 (higher is better).
A perfect pairs-F$_1$ is achieved {\em if and only if}
both the POIs and their visiting order in the
recommended trajectory are exactly the same as those in the ground truth.
On the other hand, pairs-F$_1 = 0$ means none of the recommended POI pairs was actually visited 
(in the designated order) in the real trajectory.
An illustration is shown in Figure~\ref{fig:pairf1},
the solid grey lines represent the ground truth transitions that actually visited by travellers,
and the dashed blue lines are the recommended trajectory by one of the approaches described in Section~\ref{sec:recommendation}.
Both examples have a perfect F$_1$ score, but not a perfect pairs-F$_1$ score due to the difference in POI sequencing.

\subsection{Results}
\label{sec:result}

The performance of various trajectory recommendation approaches are summarised in
Table~\ref{tab:f1} and Table~\ref{tab:pairf1},
in terms of F$_1$ and pairs-F$_1$ scores respectively.
It is apparent that algorithms 
captured information about the problem (Table~\ref{tab:algsummary})
outperform the \textsc{Random} baseline in terms of both metrics on all five datasets.

Algorithms based on POI ranking yield strong performance, in terms of both metrics, by exploring POI and query specific features.
\textsc{PoiRank} improves notably upon \textsc{PoiPopularity} and \textsc{PersTour} by leveraging more features. 
In contrast, \textsc{Markov} which leverages only POI transitions does not perform as well.
Algorithms with ranking information (\textsc{Rank+Markov} and \textsc{Rank+MarkovPath})
always outperform their respective variants with transition information alone (\textsc{Markov} and \textsc{MarkovPath}).

We can see from Table~\ref{tab:f1} that, in terms of F$_1$, \textsc{MarkovPath} and \textsc{Rank+MarkovPath}
outperform their corresponding variants \textsc{Markov} and \textsc{Rank+Markov} without the path constraints,
which demonstrates that eliminating sub-tours improves point recommendation.
This is not unexpected, 
as sub-tours worsen the proportion of correctly recommended POIs since a length constraint is used.
In contrast, most Markov chain entries have better performance in terms of pairs-F$_1$ (Table~\ref{tab:pairf1}), 
which indicates Markov chain approaches generally respect the transition patterns between POIs.

\textsc{PersTour}~\cite{ijcai15} always performs better than its variant \textsc{PersTour-L},
in terms of both metrics, especially on Glasgow and Toronto datasets.
This indicates the time budget constraint is more helpful than length constraint for recommending trajectories.
Surprisingly, we observed that \textsc{PersTour} is outperformed by \textsc{Random} baseline on Melbourne dataset. 
It turns out that on this dataset, many of the ILP problems
which \textsc{PersTour} needs to solve to get the recommendations are difficult ILP instances.
In the leave-one-out evaluation, although we utilised a large scale computing cluster with modern hardware,
$12\%$ of evaluations failed as the ILP solver was unable to find a feasible solution after $2$ hours.
Furthermore, a lot of recommendations were suboptimal solutions of the corresponding ILPs due to
the time limit. These factors lead to the inconsistent performance of \textsc{PersTour} on Melbourne dataset.

\subsection{An Illustrative Example}
\label{sec:example}

\begin{figure*}[t]
	\centering
	\includegraphics[width=\textwidth]{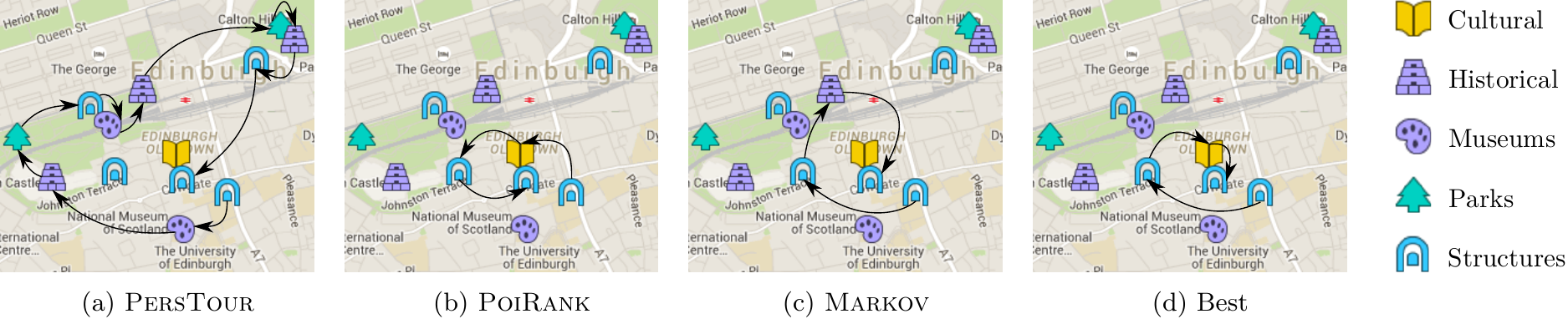}
	\caption{Different recommendations from algorithm variants.
    See the main text in Section~\ref{sec:example} for description.}
	\label{fig:exampleresult}
\end{figure*}

Figure~\ref{fig:exampleresult} illustrates an example from Edinburgh.
The ground truth is a trajectory of length $4$ that starts at a POI of category \textit{Structures},
visits two intermediate POIs of category \textit{Structures} and \textit{Cultural} and
terminates at a POI of category \textit{Structures}.
The trajectory recommended by \textsc{PersTour} is a tour with $11$ POIs, as shown in Figure~\ref{fig:exampleresult}(a),
with none of the desired intermediate POIs visited.
\textsc{PoiRank} (Figure~\ref{fig:exampleresult}(b)) recommended a tour with correct POIs,
but with completely different routes.
On the other hand, \textsc{Markov} (Figure~\ref{fig:exampleresult}(c)) missed one POI
but one of the intermediate routes is consistent with the ground truth.
The best recommendation, as shown in Figure~\ref{fig:exampleresult}(d),
with exactly the same points and routes as the ground truth,
which in this case is achieved by \textsc{Rank+MarkovPath}.

\section{Discussion and Conclusion}
\label{sec:conclusion}

In this paper, we propose an approach to recommend trajectories
by jointly optimising point preferences and routes.
This is in contrast to related work which looks at only POI or next location recommendation.
Point preferences are learned by ranking according to POI and query features,
and factorised transition probabilities between POIs
are learned from previous trajectories extracted from social media.
We investigate the maximum likelihood sequence approach (which
may recommend sub-tours) and propose an improved sequence recommendation method.
Our feature driven approach naturally allows learning the combination of POI ranks and routes.

We argue that one should measure performance with respect to the visiting order of POIs,
and suggest a new pairs-F$_1$ metric.
We empirically evaluate our tour recommendation approaches on five datasets extracted from
Flickr photos, and demonstrate that our method improves on prior work,
in terms of both the traditional F$_1$ metric and our proposed performance measure.
Our promising results from learning points and routes for trajectory recommendation suggests
that research in this domain should consider both information sources simultaneously.

\section*{Acknowledgements}
We thank Kwan Hui Lim for kindly providing his R code to reproduce his experiments.
This work is supported in part by the Australian Research Council via the Discovery Project program DP140102185.

\bibliographystyle{abbrv}
\bibliography{ref}

\onecolumn

\appendix

\urlstyle{tt}

\section{POI Features for Ranking}

\begin{table*}[ht]
\caption{Features of POI $p$ used in rankSVM given query $(p_s, p_e, L)$}
\label{tab:featurerank}
\centering
\setlength{\tabcolsep}{10pt} 
\begin{tabular}{l|l} \hline
\textbf{Feature}  & \textbf{Description} \\ \hline
\texttt{category}               & one-hot encoding of the category of $p$ \\
\texttt{neighbourhood}          & one-hot encoding of the POI cluster that $p$ resides in \\
\texttt{popularity}             & logarithm of POI popularity of $p$ \\
\texttt{nVisit}                 & logarithm of the total number of visit by all users at $p$ \\
\texttt{avgDuration}            & logarithm of the average duration at $p$ \\ \hline
\texttt{trajLen}                & trajectory length $L$, i.e., the number of POIs required \\
\texttt{sameCatStart}           & $1$ if the category of $p$ is the same as that of $p_s$, $-1$ otherwise \\
\texttt{sameCatEnd}             & $1$ if the category of $p$ is the same as that of $p_e$, $-1$ otherwise \\
\texttt{sameNeighbourhoodStart} & $1$ if $p$ resides in the same POI cluster as $p_s$, $-1$ otherwise \\
\texttt{sameNeighbourhoodEnd}   & $1$ if $p$ resides in the same POI cluster as $p_e$, $-1$ otherwise \\
\texttt{distStart}              & distance between $p$ and $p_s$, calculated using the Haversine formula \\
\texttt{distEnd}                & distance between $p$ and $p_e$, calculated using the Haversine formula \\
\texttt{diffPopStart}           & real-valued difference in POI popularity of $p$ from that of $p_s$ \\
\texttt{diffPopEnd}             & real-valued difference in POI popularity of $p$ from that of $p_e$ \\
\texttt{diffNVisitStart}        & real-valued difference in the total number of visit at $p$ from that at $p_s$ \\
\texttt{diffNVisitEnd}          & real-valued difference in the total number of visit at $p$ from that at $p_e$ \\
\texttt{diffDurationStart}      & real-valued difference in average duration at $p$ from that at $p_s$ \\
\texttt{diffDurationEnd}        & real-valued difference in average duration at $p$ from that at $p_e$ \\ \hline
\end{tabular}
\end{table*}

\begin{figure*}[ht]
	\centering
	\includegraphics[width=0.7\textwidth]{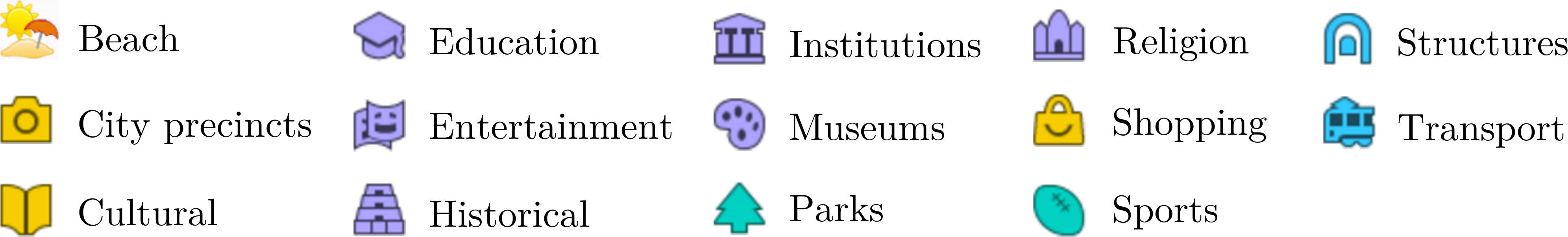}
	\caption{POI Categories}
	\label{fig:poicats}
\end{figure*}

\begin{figure*}[t]
\includegraphics[width=\textwidth]{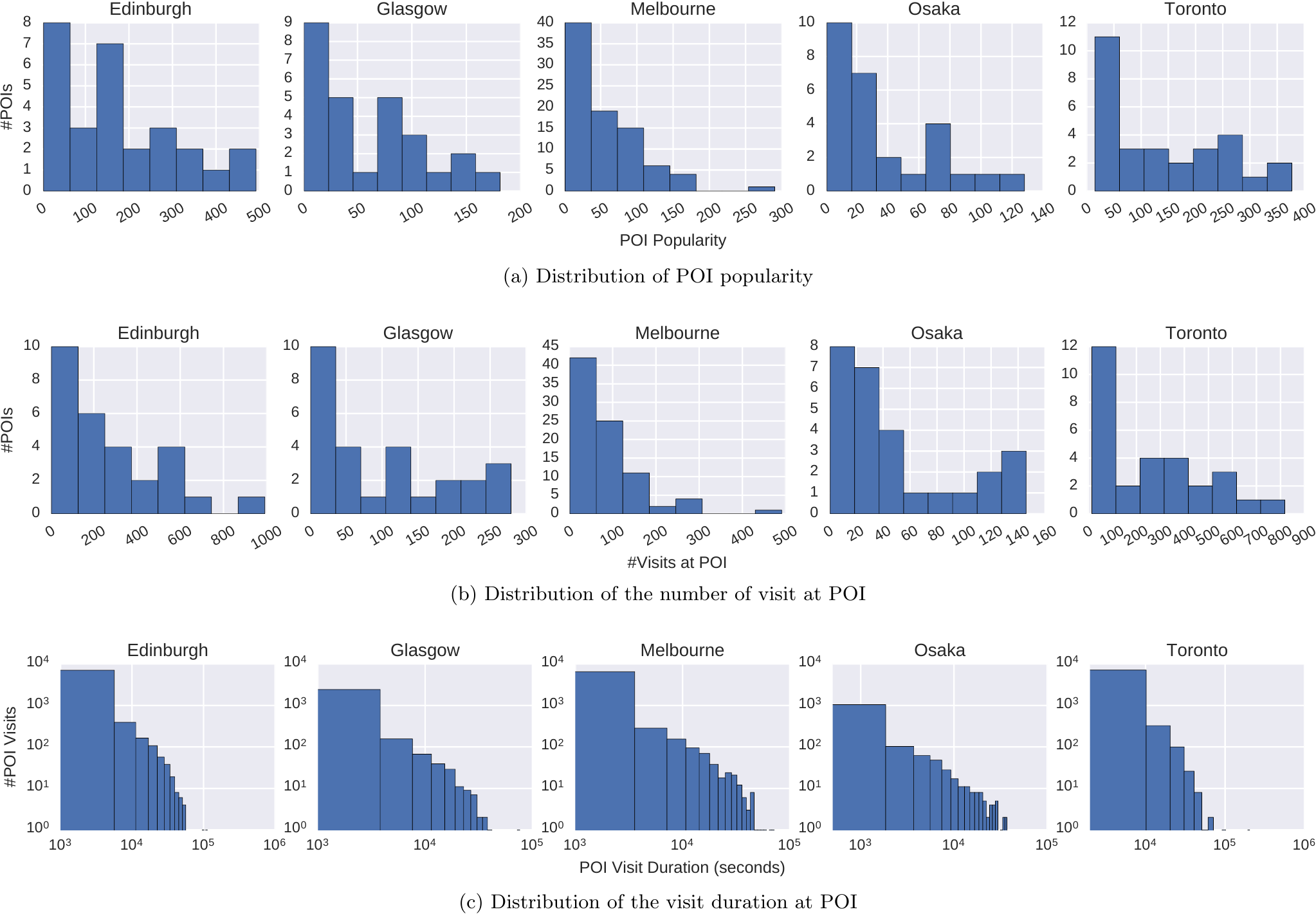}
\caption{Distribution of POI popularity, the number of visit and visit duration}
\label{fig:distro}\captionmoveup
\end{figure*}

We described an algorithm to recommend trajectories based on ranking POIs (\textsc{PoiRank}) in Section~\ref{sec:rankplan},
the features used to rank POIs are POI and query specific, as described in Table~\ref{tab:featurerank}.

Categories of POIs in all of the five trajectory datasets are show in Figure~\ref{fig:poicats}.
The distribution of POI popularity, the number of visit and average visit duration are shown in Figure~\ref{fig:distro}.

To rank POIs, features described in Table~\ref{tab:featurerank} are scaled to range $[-1.0, 1.0]$ using the same approach
as that employed by libsvm (\url{http://www.csie.ntu.edu.tw/~cjlin/libsvm/}),
i.e., fitting a linear function $f(x) = a x + b$ for feature $x$ such that the maximum value of $x$ maps to $1.0$
and the minimum value maps to $-1.0$.

\section{Transition Probabilities}

\begin{table}[ht]
\caption{POI features used to factorise POI-POI transition probabilities}
\label{tab:featuretran}
\centering
\setlength{\tabcolsep}{28pt} 
\begin{tabular}{l|l} \hline
\textbf{Feature}       & \textbf{Description} \\ \hline
\texttt{category}      & category of POI \\
\texttt{neighbourhood} & the cluster that a POI resides in \\
\texttt{popularity}    & (discretised) popularity of POI \\
\texttt{nVisit}        & (discretised) total number of visit at POI \\
\texttt{avgDuration}   & (discretised) average duration at POI \\ \hline
\end{tabular}
\end{table}

We compute the POI-POI transition matrix by factorising transition probabilities from POI $p_i$ to POI $p_j$ as a product of transition probabilities
between pairs of individual POI features, which are shown in Table~\ref{tab:featuretran}.

\begin{figure*}[htbp]
\includegraphics[width=\textwidth]{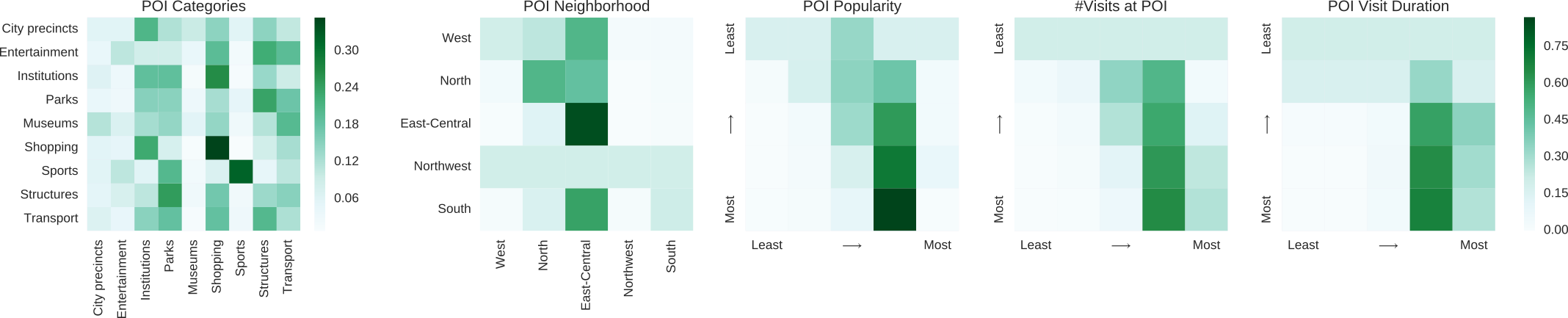}
\caption{Transition matrices for five POI features: POI category, neighbourhood, popularity, number of visits, and visit duration. These statistics are from the Melbourne dataset.}
\label{fig:transmat_all}
\end{figure*}

POI Features are discretised as described in Section~\ref{sec:feature} and transition matrices of individual features are computed using maximum likelihood estimation,
i.e., counting the number of transitions for each pair of features then normalising each row,
taking care of zeros by adding a small number $\epsilon$
\footnote{In our experiments, $\epsilon = 1$.}
to each count before normalisation.
Figure~\ref{fig:transmat_all} visualises the transition matrices for individual POI features in Melbourne.

The POI-POI transition matrix is computed by taking the Kronecker product of the transition matrices for the individual features,
and then updating it with the following constraints:

\begin{itemize}
\item 
Firstly, we disallow self transitions by setting probability of ($p_i$ to $p_i$) to zero.

\item 
Secondly, when a group of POIs have identical (discretised) features (say a group with $M$ POIs),
we distribute the probability uniformly among POIs in the group,
in particular, the incoming (unnormalised) transition probability (say, $P_{in}$) of the group computed by taking the Kronecker product is divided 
uniformly among POIs in the group (i.e., $\frac{P_{in}}{M}$), which is equivalent to choose a POI in the group uniformly at random.
Moreover, the outgoing (unnormalised) transition probability of each POI is the same as that of the group,
since in this case, \textit{the transition from any POI in the group to one outside the group represents an outgoing transition from that group}.
In addition, the self-loop transition of the group represents transitions from a POI in the group to other POIs ($M-1$ POIs) in the same group,
\textit{similar to the outgoing case}, the (unnormalised) self-loop transition probability (say $P_o$) is divided uniformly (i.e., $\frac{P_o}{M-1}$),
which corresponds to choose a transition (from $p_i$) among all transitions to the other $M-1$ POIs (exclude self-loop $p_i$ to $p_i$)
in that group uniformly at random.

\item 
Lastly, we remove feature combinations that has no POI in dataset and normalise each row of the (unnormalised) POI-POI transition matrix to form a valid probability distribution for each POI.

\end{itemize}

\section{Experiment}

\subsection{Dataset}
Trajectories used in experiment (Section~\ref{sec:experiment}) are extracted using geo-tagged photos in the Yahoo! Flickr Creative Commons 100M
(a.k.a. YFCC100M) dataset~\cite{thomee2016yfcc100m} as well as the Wikipedia web-pages of points-of-interest (POI).
Photos are mapped to POIs according to their distances calculated using the Haversine formula~\cite{haversine},
the time a user arrived a POI is approximated by the time the first photo taken by the user at that POI,
similarly, the time a user left a POI is approximated by the time the last photo taken
by the user at that POI.
Furthermore, sequence of POI visits by a specific user are divided into several pieces according to
the time gap between consecutive POI visits, and the POI visits in each piece are connected in temporal order
to form a trajectory~\cite{ht10, ijcai15}.

\subsection{Parameters}
We use a $0.5$ trade-off parameter for \textsc{PersTour} and \textsc{PersTour-L}, found to be the best weighting in~\cite{ijcai15}.
The regularisation parameter $C$ in rankSVM is $10.0$.
The trade-off parameter $\alpha$ in \textsc{Rank+Markov} and \textsc{Rank+MarkovPath} is tuned using cross validation.
In particular, we split trajectories with more than $2$ POIs in a dataset into two (roughly) equal parts,
and use the first part (i.e., validation set) to tune $\alpha$ (i.e., searching value of $\alpha$ such that \textsc{Rank+Markov} achieves the best performance on validation set, in terms of the mean of pairs-F$_1$ scores from leave-one-out cross validation),
then test on the second part (leave-one-out cross validation) using the tuned $\alpha$, and vice verse.

\subsection{Implementation}
We employ the rankSVM implementation in libsvmtools (\url{https://www.csie.ntu.edu.tw/~cjlin/libsvmtools/}).
Integer linear programming (ILP) are solved using Gurobi Optimizer (\url{http://www.gurobi.com/})
and lp\_solve (\url{http://lpsolve.sourceforge.net/}).
Dataset and code for this work are available in repository \url{https://bitbucket.org/d-chen/tour-cikm16}.

\subsection{Performance metric}
A commonly used metric for evaluating POI and trajectory recommendation is
the F$_1$ score on points~\cite{ijcai15},
Let $\mathcal{T}$ be the trajectory that was visited in the real world,
and $\hat{\cal T}$ be the recommended trajectory,
$\mathcal{P}_{\mathcal{T}}$ be the set of POIs visited in $\mathcal{T}$,
and $\mathcal{P}_{\hat{\mathcal{T}}}$ be the set of POIs visited in $\hat{\mathcal{T}}$,
F$_1$ score on points is the harmonic mean of precision and recall of POIs in trajectory,
\begin{equation*}
F_1= \frac{2  P_{\textsc{point}}  R_{\textsc{point}}}
          {P_{\textsc{point}} + R_{\textsc{point}}}, 
\text{~where~}
P_{\textsc{point}} = \frac{|\mathcal{P}_{\mathcal{T}} \cap \mathcal{P}_{\hat{\mathcal{T}}}|}
                          {|\hat{\mathcal{T}}|}
\text{~and~}
R_{\textsc{point}} = \frac{|\mathcal{P}_{\mathcal{T}} \cap \mathcal{P}_{\hat{\mathcal{T}}}|}
                          {|\mathcal{T}|}.
\end{equation*}

A perfect F$_1$ (i.e., F$_1 = 1$) means the POIs in
the recommended trajectory are exactly the same set of POIs as those in the ground truth,
and F$_1 = 0$ means that none of the POIs in the real trajectory was recommended.

While F$_1$ score on points is good at measuring whether POIs are correctly recommended,
it ignores the visiting order between POIs.
$\text{Pairs-F}_1$ takes into account both the point identity and the visiting orders in a trajectory.
This is done by measuring the F$_1$ score of every pair of ordered POIs, whether they are adjacent or not in trajectory,
\begin{equation*}
\text{pairs-F}_1 = \frac{2 P_{\textsc{pair}} R_{\textsc{pair}}}
                        {P_{\textsc{pair}} + R_{\textsc{pair}}},
\text{~where~}
P_{\textsc{pair}} = \frac{N_c} {|\hat{\mathcal{T}}|(|\hat{\mathcal{T}}|-1) / 2} \text{~and~}
R_{\textsc{pair}} = \frac{N_c} {|\mathcal{T}|(|\mathcal{T}|-1) / 2},
\end{equation*}
and $N_c$
\footnote{We define pairs-F$_1 = 0$ when $N_c = 0$.}
is the number of ordered POI pairs $(p_j, p_k)$ that
appear in both the ground-truth and the recommended trajectories,
\begin{align*}
    (p_j \prec_{\mathcal{T}} p_k ~\land~ p_j \prec_{\hat{\mathcal{T}}} p_k)  ~\lor~
    (p_j \succ_{\mathcal{T}} p_k ~\land~ p_j \succ_{\hat{\mathcal{T}}} p_k),
\end{align*}
with $p_j \ne p_k, ~p_j, p_k \in \mathcal{P}_{\mathcal{T}} \cap \mathcal{P}_{\hat{\mathcal{T}}}, ~1 \le j \ne k \le |\mathcal{T}|$.
Here $p_j \prec_{\mathcal{T}} p_k$ denotes POI $p_j$ was visited before POI $p_k$ in trajectory $\mathcal{T}$
and $p_j \succ_{\mathcal{T}} p_k$ denotes $p_j$ was visited after $p_k$ in $\mathcal{T}$.

Pairs-F$_1$ takes values between 0 and 1. A perfect pairs-F$_1$ (1.0) is achieved if and only if
both the POIs and their visiting orders in the recommended trajectory are exactly the same as those in the ground truth.
Pairs-F$_1 = 0$ means none of the recommended POI pairs was actually visited in the real trajectory.

Performance data reported in Table~\ref{tab:f1} and Table~\ref{tab:pairf1} are the mean and standard deviation of instances successfully recommended by
all methods shown in Table~\ref{tab:algsummary}.

\end{document}